\documentclass[10pt,twocolumn,letterpaper]{article}

\usepackage{iccv}
\usepackage[accsupp]{axessibility}
\usepackage{times}
\usepackage{epsfig}
\usepackage{graphicx}
\usepackage{amsmath}
\usepackage{amssymb}
\usepackage{savesym}
\savesymbol{checkmark}
\usepackage{booktabs}
\usepackage{tabularx} % in the preamble
\usepackage{enumitem}
\usepackage{algorithm}
\usepackage{dingbat}

\usepackage{mwe}
\long\def\/*#1*/{}
\usepackage{dsfont}
\usepackage{microtype}
\usepackage{cite}
\usepackage{color, colortbl}

\usepackage{comment}
\usepackage{multirow}
\usepackage{adjustbox}
\usepackage{wrapfig}
\usepackage{subcaption}
\usepackage[hang,flushmargin]{footmisc}
\definecolor{Gray}{gray}{0.9}

\usepackage[pagebackref=true,breaklinks=true,letterpaper=true,colorlinks,bookmarks=false]{hyperref}
\usepackage[capitalize]{cleveref}
\crefname{section}{Sec.}{Secs.}
\Crefname{section}{Section}{Sections}
\Crefname{table}{Table}{Tables}
\crefname{table}{Tab.}{Tabs.}
\Crefname{tab}{Table}{Tables}
\crefname{tab}{Tab.}{Tabs.}
\Crefname{figure}{Figure}{Figures}
\crefname{figure}{Fig.}{Figs.}

\iccvfinalcopy % *** Uncomment this line for the final submission

\def\iccvPaperID{5547} % *** Enter the ICCV Paper ID here

\def \OURMETHOD {MGM\xspace}

\begin{document}

%%%%%%%%% TITLE
\title{Motion-Guided Masking for Spatiotemporal Representation Learning}

\author{David Fan$^1$ \qquad Jue Wang$^1$ \qquad Shuai Liao$^2$ \qquad Yi Zhu$^3$ \qquad Vimal Bhat$^1$ \\ Hector Santos-Villalobos$^1$ \qquad Rohith MV$^1$ \qquad Xinyu Li$^1$ \\
$^1$ Amazon Prime Video \qquad $^2$ Amazon Fulfillment Technology \qquad $^3$ AWS AI Labs \\
{\tt\small \{fandavi, juewangn, uliaoshu, yzaws, vimalb, hsantosv, kurohith, xxnl\}@amazon.com}
}

% \and

\maketitle

\begin{abstract}

\noindent Several recent works have directly extended the image masked autoencoder (MAE) with random masking into video domain, achieving promising results. However, unlike images, both spatial and temporal information are important for video understanding. This suggests that the random masking strategy that is inherited from the image MAE is less effective for video MAE. This motivates the design of a novel masking algorithm that can more efficiently make use of video saliency. Specifically, we propose a motion-guided masking algorithm (\OURMETHOD{}) which leverages motion vectors to guide the position of each mask over time. Crucially, these motion-based correspondences can be directly obtained from information stored in the compressed format of the video, which makes our method efficient and scalable. On two challenging large-scale video benchmarks (Kinetics-400 and Something-Something V2), we equip video MAE with our \OURMETHOD{} and achieve up to +$1.3\%$ improvement compared to previous state-of-the-art methods. Additionally, our \OURMETHOD{} achieves equivalent performance to previous video MAE using up to $66\%$ fewer training epochs. Lastly, we show that \OURMETHOD{} generalizes better to downstream transfer learning and domain adaptation tasks on the UCF101, HMDB51, and Diving48 datasets, achieving up to +$4.9\%$ improvement compared to baseline methods.
\end{abstract}
\section{Introduction}
\label{sec:introduction}
\noindent Video transformers~\cite{arnab2021vivit, liu2021swin, li2022mvitv2, wang2022deformable, patrick2021keeping} have achieved state-of-the-art for a variety of video understanding tasks, mirroring the success of image transformers such as ViT~\cite{dosovitskiy2020image}. However, many video transformers heavily rely on large-scale supervised pretraining from image datasets such as ImageNet21K~\cite{deng2009imagenet} and JFT-300M~\cite{sun2017revisiting}, which is data-inefficient.

\begin{figure}[t]
    \centering
    \includegraphics[width=0.49\textwidth]{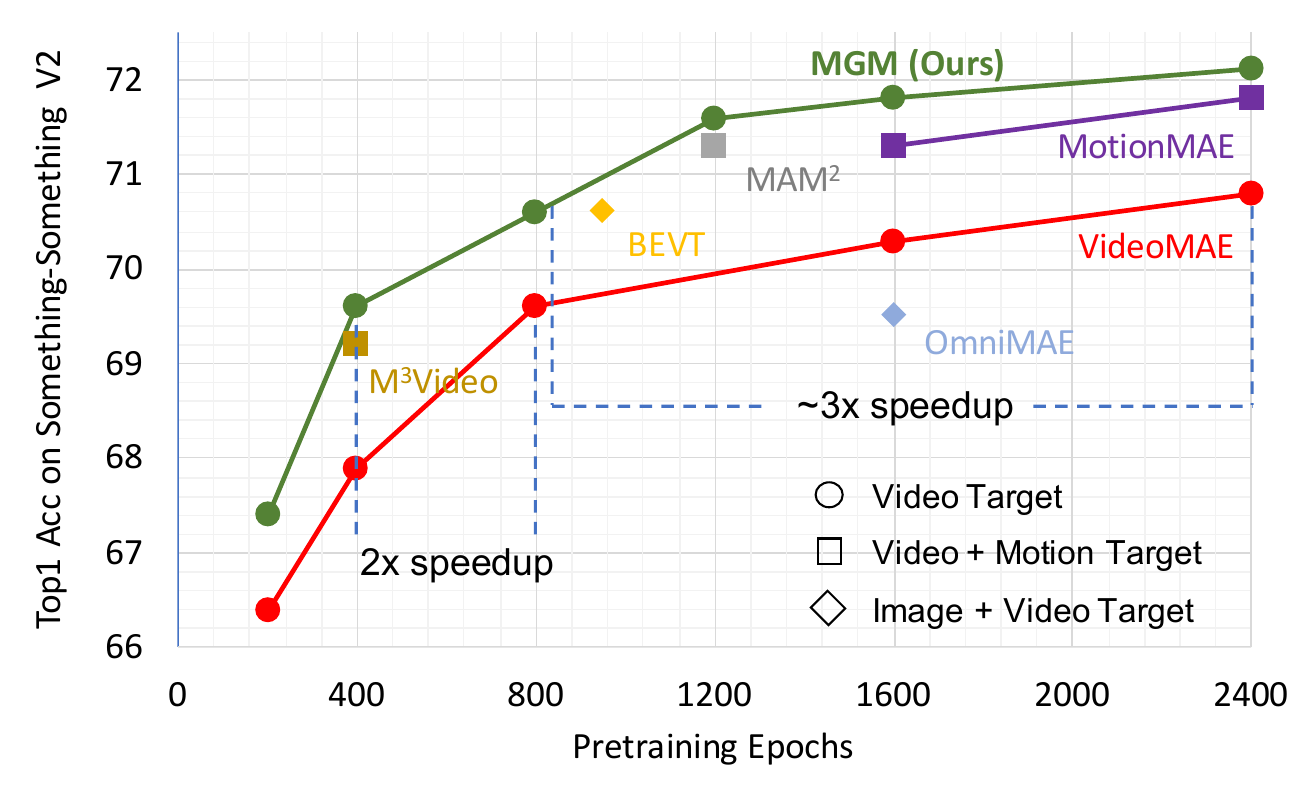}
    \caption{Top-1 accuracy on Something-Something V2. Our proposed \OURMETHOD{} outperforms all previous masked-video methods (M$^{3}$Video~\cite{Sun2022M3VideoMM}, MAM$^{2}$~\cite{song2022takes}, BEVT ~\cite{Wang2022BEVTBP}, OmniMAE~\cite{Girdhar2022OmniMAESM}, MotionMAE~\cite{yang2022self}, and VideoMAE~\cite{Tong2022VideoMAEMA}) with improved training efficiency. Marker shape denotes reconstruction target.}
    \label{fig:teaser}
    \vspace{-5mm}
\end{figure}

Self-supervised learning (SSL)~\cite{grill2020bootstrap, wang2022long, chen2020simple, He2020MomentumCF, Recasens2021BroadenYV} is one promising paradigm for eliminating the dependency on large-scale supervised pretraining. The denoising / masked autoencoder (MAE)~\cite{vincent2008extracting}, which was originally popularized by BERT~\cite{kenton2019bert} for natural language modeling, has recently re-emerged as a promising representation learning method for vision tasks. Image MAE~\cite{he2022masked, bao2021beit} has achieved state-of-the-art in image domain via learning to reconstruct randomly masked image patches. Recently, similar works~\cite{Tong2022VideoMAEMA, Feichtenhofer2022MaskedAA} that reconstruct randomly masked video have obtained encouraging results in video domain. However, few SSL works focus on effectively learning video saliency. In this paper, we improve upon the previous video MAEs~\cite{Tong2022VideoMAEMA, Feichtenhofer2022MaskedAA} via video saliency.

We argue that the random masking strategy which is inherited from image MAE is not optimal for video. The optimal random masking ratio for video MAE~\cite{Tong2022VideoMAEMA, Feichtenhofer2022MaskedAA} is higher than that for image MAE~\cite{he2022masked} (0.9 vs. 0.75). This can be understood as a consequence of the natural temporal coherence within videos, which leads to the existence of similar video patches in other frames. When using random masking, many of these correlated video patches may be visible to the encoder which would make reconstruction easier. This necessitates the use of a high masking ratio in order to reduce redundancy and make the reconstruction task sufficiently challenging~\cite{Feichtenhofer2022MaskedAA}. However, increasing the masking ratio has the side-effect of leaving fewer visible patches for the MAE encoder to learn spatiotemporal saliency from, which we hypothesize limits the learned representation.

We thus propose to guide the model to learn to reconstruct the most salient regions of video. As humans and objects are key to understanding video, one natural idea is to track the bounding boxes in each video frame and mask the content within. This is also consistent with the observation that video pixels evolve continuously frame-by-frame, and therefore, the MAE should be able to reconstruct this spatiotemporal continuity. However, generating bounding boxes for each frame is impractical for large-scale video datasets.

To this end, we hypothesize that motion is an effective guide for detecting spatiotemporal saliency. To test this hypothesis, we define the saliency score as $r_{bbox}$ / $r_{non\_bbox}$, or the ratio of average motion magnitude within bounding boxes to average motion magnitude outside bounding boxes. The saliency score is 1.47 for Something-Something v2~\cite{goyal2017something} and 1.28 for Kinetics-400~\cite{kay2017kinetics} respectively, suggesting that regions of higher motion overlap with spatially salient regions more often than regions of lesser motion.

We thus propose an improved masking algorithm, \OURMETHOD{}, which uses motion to continuously guide the mask to cover the most salient spatiotemporal regions. \OURMETHOD{} obtains cheap motion correspondence by exploiting the H.264 codec~\cite{richardson2002video} which prevails in popular video formats such as MP4. During the encoding phase of H.264, motion vectors are precomputed and stored as part of the codec. Thus during the decoding phase, motion vectors can be obtained ``for free'' along with RGB frames. The use of readily available motion vectors instead of expensive optical flow enables us to efficiently achieve scale. While some supervised works have used motion vectors~\cite{wu2018compressed, wang2022deformable, korbar2019scsampler}, we are the first to use motion vectors for MAE pretraining.

Our \OURMETHOD{} outperforms previous VideoMAE~\cite{Tong2022VideoMAEMA} by 1.2\% on Something-Something V2 (SSv2)~\cite{goyal2017something} and 0.2\% on Kinetics-400 (K400)~\cite{kay2017kinetics}, demonstrating that our \OURMETHOD{} is effective. \OURMETHOD{} can also achieve the same performance as~\cite{Tong2022VideoMAEMA} with 50\% fewer training epochs as shown in~\cref{fig:teaser}, further making MAE pretraining more data-efficient. We also show that \OURMETHOD{} generalizes well to small datasets (UCF101~\cite{soomro2012ucf101}, HMDB51~\cite{kuehne2011hmdb}, and Diving48~\cite{li2018resound}) in both full finetune and linear probe evaluation as well as domain adaptation settings with up to 4.9\% performance improvement over VideoMAE~\cite{Tong2022VideoMAEMA}. This shows that the features learned by \OURMETHOD{} contain richer semantics that transfer well to video recognition tasks.
% contributions
In summary, our contributions are:
\begin{enumerate}[itemsep=0pt,parsep=0pt]
    \item \OURMETHOD{}, an efficient and effective self-supervised algorithm for 3D masking that continuously models motion trajectories.
    \item Applying motion vectors which are directly available during video decoding - unlike optical flow - to provide efficient motion guidance in the MAE framework.
    \item New state-of-the-art or comparable results on two large-scale datasets and various downstream tasks on three small datasets, as well as detailed ablations and insights.
\end{enumerate}
\section{Related Works}
\label{sec:related_works}

\noindent \textbf{Image Masked Auto-Encoders.}\quad
iGPT~\cite{chen2020generative} was one of the first recent attempts to revisit masked image modeling using Transformer architecture rather than CNNs. iGPT reconstructs at the pixel-level. Other recent works try different reconstruction targets. Some works such as BEIT~\cite{bao2021beit} explore using tokenization as a prediction target, such as tokens from pretrained dVAE~\cite{van2017neural}. MaskFeat~\cite{Wei2022MaskedFP} explores using HoG features~\cite{dalal2005histograms} as a reconstruction target. ImageMAE~\cite{he2022masked} reconstructs normalized image patches and focuses on studying decoder design, discovering that larger decoders enable the encoder to process only visible patches, thus enabling the use of a larger masking ratio to achieve both better performance and computational efficiency. Unlike these works, our \OURMETHOD{} is designed specifically for video.

\noindent \textbf{Video Masked Auto-Encoders.}\quad
Some works such as VIMPAC~\cite{tan2021vimpac} and BEVT~\cite{Wang2022BEVTBP} try to reconstruct video tokens and thus require strong tokenizers using either extra pretraining data or domain-specific knowledge. VideoMAE~\cite{Tong2022VideoMAEMA, Feichtenhofer2022MaskedAA} bypasses this requirement by applying image MAE~\cite{he2022masked} to reconstruct 3D RGB video patches, achieving promising results on video benchmarks. Other works such as OmniMAE~\cite{Girdhar2022OmniMAESM} use both image and video masked modeling to achieve improved performance on both image and video tasks. A line of very recent work tries to improve VideoMAE by designing explicit motion cues by reconstructing optical flow or frame difference based targets~\cite{song2022takes, Sun2022M3VideoMM, yang2022self}.

These existing video MAE works all inherit random masking from image MAE~\cite{he2022masked} where the mask is either discontinuous such as in ST-MAE~\cite{Feichtenhofer2022MaskedAA} or random tubes such as in VideoMAE~\cite{Tong2022VideoMAEMA}. In the former, the mask is randomly distributed both spatially and temporally, and in the latter, the mask is randomly distributed spatially but temporally static.  In contrast, we re-examine this assumption and design a new masking algorithm that continuously masks out video motion to force the model to focus on spatiotemporal saliency. We make no changes to the reconstruction target nor model architecture.

\noindent \textbf{VideoMAE}~\cite{Tong2022VideoMAEMA} initially generates a random mask and statically propagates the mask across time. The position of the mask is the same across frames, forming small tubelets which are spatially discontinuous but temporally continuous.

\noindent \textbf{ST-MAE}~\cite{Feichtenhofer2022MaskedAA} independently generates a random mask per-frame which is neither spatially nor temporally continuous.

\section{Methodology}
\begin{figure*}[t]
    \centering
    \includegraphics[width=.99\linewidth]{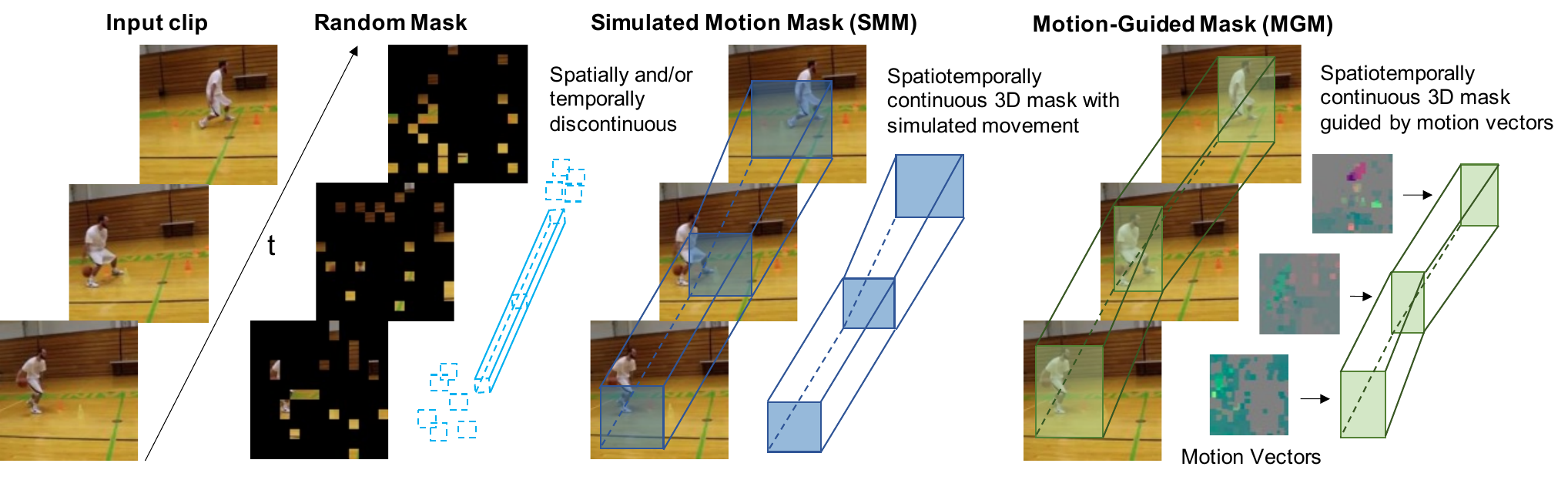}
    \caption{An overview of a) the proof-of-concept simulated-motion masking (SMM) vs. b) motion-guided masking (\OURMETHOD{}). Our \OURMETHOD{} produces 3D masks that achieves both spatial and temporal continuity, while capturing the true motion information in the video. Random masking (left-most) is both spatially and temporally discontinuous.}
    \label{fig:masking_diagram}
\end{figure*}

\noindent As detailed in~\cref{sec:introduction}, motion is concentrated in salient video regions. Random masking does not effectively cover these regions because motion is not uniformly distributed throughout the video. Furthermore, spatially salient entities such as people and objects are continuous rather than disjoint patches. In general, this leads us to believe random masking may miss certain semantics that are useful for learning spatiotemporal representations. The goal of this work is to address these deficiencies to make masked video modeling more effective for video representation learning. Because video is temporally cohesive, the MAE should be able to reconstruct a continuous spatiotemporal volume comprised of the salient regions. This motivates our following approach.

\subsection{Preliminaries}
\noindent Given a video $V \in \mathbb{R}^{T \times H \times W \times C }$, where $T,H,W,C$ denote the number of frames, height, width, and RGB-channels, the video is typically first split into patches of size $t \times h \times w \times C$ and processed with a patch embedding layer $\mathcal{P}$ to obtain a sequence of token embeddings $V_p$.

\begin{equation}
    V_p = \mathcal{P}(V); V_p \in \mathbb{R}^{\frac{T}{t} \times \frac{H}{h} \times \frac{W}{w}}
    \label{equ:patch_embedding}
\end{equation}

Next, a masking function $\eta$ (e.g. random masking~\cite{Tong2022VideoMAEMA, Feichtenhofer2022MaskedAA}) generates a binary mask $M$ which is applied to select a set of visible patches with mask ratio $\gamma$.

\begin{equation}
\begin{split}
    M &= \eta(V_p,\gamma); \; dim(M) = dim(V_p) \\
    V_{\text{p\_visible}} &= V_p \cdot (\sim M) \\
    V_{\text{p\_masked}} &= V_p \cdot M % should be mask token not the actual token?
\end{split}
\label{equ:masking_tokens}
\end{equation}

The encoder $\phi$ then processes only the visible patches $V_{\text{p\_visible}}$ while the decoder $\xi$ processes the full set of encoded patches and masked tokens $\phi(V_{\text{p\_visible}}) \cup V_{\text{p\_masked}}$ to reconstruct the video. This work uses the same asymmetric encoder-decoder design as~\cite{Tong2022VideoMAEMA, Feichtenhofer2022MaskedAA, he2022masked}.

\begin{equation}
    E=\phi(V_{\text{p\_visible}})
    \label{equ:mae_encoder}
\end{equation}
\begin{equation}
    V'=\xi(E \cup V_{\text{p\_masked}})
    \label{equ:mae_decoder}
\end{equation}

Finally, the model is trained with MSE reconstruction loss between $V$ and $V'$. The model is then transferred to downstream tasks such as classification via finetuning with cross-entropy loss. Note that this work focuses on the mask generator $\eta(\gamma)$, which is complementary to recent works which look into the loss function and reconstruction targets~\cite{song2022takes, Wang2022BEVTBP, Sun2022M3VideoMM, yang2022self}.

\subsection{Simulated-Motion Masking (SMM)}
\label{sec:methods:simulated_motion_masking}
\noindent To help the encoder learn spatiotemporal semantics throughout the video, we propose to generate a spatiotemporally continuous moving 3D mask, as shown in~\cref{fig:masking_diagram}. 
% In this way, higher level learning is required to reconstruct the masked parts. 
This differs from random masking~\cite{Feichtenhofer2022MaskedAA} and random tube masking~\cite{Tong2022VideoMAEMA} which break spatiotemporal continuity.

We define the mask as a vector $[x, y, w, h]$ where $x$ and $y$ define the top-left coordinate and $w$ and $h$ define the width and height of the mask. We first explore the use of random motion to generate a dense moving 3D mask with masking ratio $\gamma$. To do this, we initialize a rectangle shaped mask in the first frame with dimensions $w_0 = h_0 = \sqrt{\gamma * \frac{H}{h} * \frac{W}{w}}$. $x_0$ and $y_0$ are then randomly initialized in $rand(0, \frac{H}{h} - h_0)$. We then propagate the mask across time using the following recurrent relation:
%In this work, each $16 \times 224 \times 224$ video is split into $2 \times 16 \times 16$ patches following~\cite{Tong2022VideoMAEMA}, so with a masking ratio $\gamma=0.75$,  $w_0 = h_0 = 12$
\begin{equation}
\begin{split}
    % [x_t,y_t,w,h]=[x_t+v_x,y_t+v_y,w*s_t,h*s_t], t\in[1,T)
    [x_t, y_t, w_t, h_t] = &[x_{t-1} + v_{xt}, y_{t-1} + v_{yt}, \\&w_{t-1} + s_t, h_{t-1} + s_t] \\
    t \in [1,T)&, 
    \sum_{t}{s_{t}} = 0
\end{split}
\label{equ:random_masking}
\end{equation}
% where for frame $t$, $x_{t}$ and $y_{t}$ are the top-left coordinates of the mask, $w_t$ and $y_t$ are the width and height of the mask,
where $v_{xt}, v_{yt}$ are the random horizontal and vertical velocity components, and $s_t$ is a random scaling factor. To ensure that the masking ratio for each video is consistent, we enforce that $\sum_{t}{s_t} = 0$; in other words, the number of masked patches remains fixed for a given $\gamma$.

\subsection{Motion-Guided Masking (\OURMETHOD{})}
\label{sec:methods:guided_motion_masking}
\noindent While SMM leads to an improvement over random masking, SMM generates masks that may not reflect true motion as it is not context-aware. For example in~\Cref{fig:masking_diagram}, the mask may cover static background rather than the basketball player depending on the random initialization and velocity. The most straightforward approach to address this issue would be to guide the mask using precomputed object bounding boxes or optical flow. However, this requires additional data and cost and would make such a method less scalable.

We propose to use motion vectors as guidance to ensure that the mask movement consistently covers motion across frames. Motion vectors~\cite{richardson2002video, wu2018compressed} can be decoded from raw video (with H.264 or H.265 codecs) with no extra cost, which ensures that our proposed approach adds no additional cost and scales well to larger datasets. A motion vector $M_{t}[x,y]$ for a given pixel position $x, y$ at frame $t$, is defined as $[d_x, d_y]$ where $d_x$ and $d_y$ denote the displacement of pixel $x, y$ from frame $t-1$. Motion vectors are typically sparsely assigned to $8 \times 8$ pixel grids~\cite{richardson2002video, wu2018compressed}, so the overall motion vector map is similar to optical flow but at a lower resolution. In order to use motion vectors to guide the masking, we need pixel-level correspondence between the motion vectors and the video. We accomplish this by upsampling the motion vector map to match the spatial resolution of the video.

Given a motion vector map $M_{t}\in \mathbb{R}^{T\times \frac{H}{8}\times \frac{W}{8}}$, we first use nearest-neighbor upsampling ($U_\text{nearest}$) to get $M_{\text{t\_scaled}} \in \mathbb{R}^{T \times H \times W}$.
% \vspace{-0.25cm}
\begin{equation}
% \vspace{-0.2cm}
    \text{M}_{\text{t\_scaled}} = U_\text{nearest}(M_t, (H, W))
\end{equation}
\noindent Next, we initialize a dense rectangle mask for frame $t_0$ with dimensions $w_0 = h_0 = \sqrt{\gamma * \frac{H}{h} * \frac{W}{w}}$. $x_0$ and $y_0$ are then randomly initialized in $rand(0, \frac{H}{h} - h_0)$. The mask is then generated for following time-steps as:
\begin{equation}
\begin{split}
    x_{\text{t\_center}}, y_{\text{t\_center}} &= Pool_\text{topK}(M_{\text{t\_scaled}}) \\
    w_t &= w_{t-1} + s_{xt} \\
    h_t &= h_{t-1} + s_{yt} \\
    % w_t &= \sqrt{\gamma * \frac{H}{h} * \frac{W}{w}} + s_{xt} \\
    % h_t &= \sqrt{\gamma * \frac{H}{h} * \frac{W}{w}} + s_{yt} \\
    [x_t, y_t, w_t, h_t] &= [x_{\text{t\_center}} - \frac{w_t}{2}, y_{\text{t\_center}} - \frac{y_t}{2}, w_t, h_t] \\
    t \in [1,T),
    \sum_{t}&{s_{xt}} = 0, \sum_{t}{s_{yt}} = 0
\end{split}
\label{equ:mgm_masking}
\end{equation}
% \begin{equation}
%     \text{Mask}_t=U_\text{nearest}(Pool_\text{topK}(M_t,\gamma))
% \end{equation}
\noindent where $Pool_\text{topK}$ is the topK (top-1 in our work) magnitude pooling operation that gives the argmax indexes $x_{\text{t\_center}}, y_{\text{t\_center}}$, which the mask is centered around. We also apply jitter $s_{xt}$, $s_{yt}$ to both the mask height and width.
%To ensure a consistent batch size, we truncate each mask to match the length of the shortest mask in the batch.

In other words, \OURMETHOD{} produces a moving 3D mask where the movement of the mask is centered around parts of the video with the highest magnitude of motion, as shown in~\cref{fig:masking_diagram}. This makes the reconstruction task more focused on spatiotemporal semantics than random masking.
\section{Experimental Results}
\label{sec:results}
\begin{table*}[!t]
\small
\begin{center}
    \begin{tabular}{llcclcclc}
     &&& \multicolumn{3}{c}{\textit{\textbf{finetune eval on SSv2}}} & \multicolumn{3}{c}{\textit{\textbf{finetune eval on K400}}}   \\ 
		\cmidrule(lr){4-6} \cmidrule(lr){7-9} 
    \textbf{Model} & \textbf{Backbone} & \textbf{GFLOPs} & \textbf{Epochs} & \textbf{Pretrain} & \textbf{Top1} & \textbf{Epochs} & \textbf{Pretrain} & \textbf{Top1}  \\
    \hline
    \multicolumn{7}{l}{\textbf{Supervised}} \\
    SlowFast~\cite{feichtenhofer2019slowfast}     & R101 + NL        & $106$    & 196 & K400           & 63.1   & - & -   &  79.8  \\
    TimeSformer~\cite{bertasius2021space}  & ViT-B            &$196$     & 15 & IN21K           &59.5   & 15 & IN21K   & 78.0   \\
    VidTr~\cite{zhang2021vidtr}        & ViT-B            &$351$     & 50 & K400           & 63.0   & - & -   & 79.1   \\
    MotionFormer~\cite{patrick2021keeping} & ViT-B            &$370$     & 90+35 & IN21K + K400          & 66.5   & 90 & IN21K   & 79.7   \\
    MViTv1~\cite{fan2021multiscale} & MViT-B & $455$    & 200 & K400           &67.7   & - & -   & 81.2   \\ % 64 x 3
    MViTv2~\cite{li2022mvitv2} & MViT-B & $225$    & 200 & K400           & 70.5   & - & -   & 82.9   \\ % 32 x 3
    \hline      \multicolumn{7}{l}{\textbf{Self-supervised}} \\
    BEVT~\cite{Wang2022BEVTBP} & Swin-B & 282 & 800+150 & IN1K + K400 & 70.6 & 800+150 & IN1K + K400 & 80.6 \\
    M$^{3}$Video~\cite{Sun2022M3VideoMM}    & ViT-B            &$180$  & 400    & SSv2  & 69.2  & 400 & K400 & 79.7 \\
    % MAM${^2}$~\cite{song2022takes}       & ViT-B+ VQGAN     &$180$  & 800    & SSv2   & 70.7  & K400  & 82.3  \\
    \rowcolor{Gray}
    \OURMETHOD{}    & ViT-B            &$180$  & 400    & SSv2  & 69.6  & 400 & K400 & 80.3 \\
    VideoMAE~\cite{Tong2022VideoMAEMA}        & ViT-B            &$180$  & 800    & SSv2   & 69.6  & 800 & K400  & 80.0  \\
    \rowcolor{Gray}
    \OURMETHOD{}    & ViT-B            &$180$  & 800    & SSv2   & 70.6  & 800 & K400  & 80.8  \\
    MAM${^2}$~\cite{song2022takes}       & ViT-B + VQGAN     &$180$  & 1200   & SSv2   & 71.3  & 800 & K400  & 82.3  \\
    \rowcolor{Gray}
    \OURMETHOD{}    & ViT-B            &$180$  & 1200    & SSv2   & 71.6  & 1200 & K400  & 81.2  \\
    OmniMAE~\cite{Girdhar2022OmniMAESM} & ViT-B & 180 & 1600 & IN1K + SSv2 & 69.5 & 1600 & IN1K + K400 & 80.8 \\
    VideoMAE~\cite{Tong2022VideoMAEMA}        & ViT-B            &$180$  & 1600   & SSv2   &    70.3   & 1600 & K400  & 81.5 \\
    VideoMAE~\cite{Tong2022VideoMAEMA}        & ViT-B            &$180$  & 2400   & SSv2   & 70.8  & - & -  & - \\
    STMAE~\cite{Feichtenhofer2022MaskedAA}           & ViT-B            &$180$  & 1600   & K400   & N/A    & 1600 & K400  & 81.3    \\
    MotionMAE~\cite{yang2022self}       & ViT-B            &$180$  & 2400   & SSv2   & 71.8  & 1600 & K400  & 81.7 \\
    \rowcolor{Gray}
    \OURMETHOD{}    & ViT-B            &$180$  & 1600   & SSv2   & 71.8  & 1600 & K400  & 81.7  \\
    \rowcolor{Gray}
    \OURMETHOD{}    & ViT-B            &$180$  & 2400   & SSv2   & 72.1  & -  & - & - \\

  \end{tabular}
\end{center}
\caption{Results comparison on Something-SomethingV2 (SSv2) and Kinetics-400 (K400). GFLOPs is listed as a single view’s. All results are with fine-tuning. We apply repeated augmentation~\cite{hoffer2020augment} = 2 for this table to compare against SOTA.}
\label{table:main_res}
\end{table*}

\subsection{Datasets}
\noindent We conduct experiments on five commonly used datasets:
\noindent\textbf{Something-Something V2 (SSv2)}~\cite{goyal2017something} contains 220K videos with 174 action classes. SSv2 is considered a motion heavy dataset, as most of the labels are defined by the motion and directionality of the actual action.
% hus strong performance on SSv2 implies motion understanding.
\textbf{Kinetics-400 (K400)}~\cite{kay2017kinetics} is the de-facto standard dataset used to evaluate video recognition. It contains 240K Internet videos with 400 action classes.
\textbf{UCF101}~\cite{soomro2012ucf101} is a dataset containing 13K Internet short videos with 101 action classes.
\textbf{HMDB51}~\cite{kuehne2011hmdb} is a dataset containing 5K short movie clips from 51 action classes.
\textbf{Diving48}~\cite{li2018resound} contains 18K untrimmed video clips from 48 action classes, all of which are types of dives. We report the top-1 accuracy on the evaluation set for all datasets following standard practices~\cite{feichtenhofer2019slowfast}. Only UCF101 and HMDB51 have multiple split versions; we use split 1.
% Because the high-level action is always diving and each diving type is differentiated by specific movements, this dataset tests fine-grained motion understanding -- similar to SSv2.

\subsection{Implementation Details}

\noindent\textbf{Model Configuration:}
We experiment with both 2D and 3D transformer backbones. The default backbone is ViT-Base~\cite{dosovitskiy2020image} with global joint space-time attention. We also experimented with TimesFormer~\cite{bertasius2021space} with divided space-time attention. For fair comparison, we use the same input patch size of $2 \times 16 \times 16$ for all models following~\cite{Tong2022VideoMAEMA}.

\noindent\textbf{Pre-Processing:} 
We pretrain with clips of 16 frames sampled at a temporal stride of 4 and 2 for K400 and SSv2 respectively following~\cite{Tong2022VideoMAEMA}. We use a fixed spatial resolution of $224 \times 224$ for all experiments. We apply multi-scale-crop and horizontal flip augmentation by default (flip is not applied to SSv2). We follow~\cite{Tong2022VideoMAEMA} to use AdamW~\cite{loshchilov2018decoupled} optimizer with a base learning rate $1.5e-4$, weight decay of $0.05$, $\beta=[0.9, 0.95]$, and cosine learning rate decay. 
% Following~\cite{Tong2022VideoMAEMA}, no other data augmentations are used.
% , unlike other self-supervised works which rely on heavy color jittering.

\noindent\textbf{Finetuning and Evaluation:}
We use the same 16-frame clip for finetuning and multi-view evaluation protocol following standard practice~\cite{feichtenhofer2019slowfast}. We use TSN-style sampling~\cite{wang2018temporal} on SSv2 dataset with 2 temporal $\times$ 3 spatial views during test-time following~\cite{Tong2022VideoMAEMA} for fair comparison. For Kinetics-400, UCF101, HMDB51 and Diving48, we use 5 temporal $\times$ 3 spatial views during test-time following~\cite{Tong2022VideoMAEMA} for fair comparison. See the supplementary material for  hyperparameter details which are mostly the same as~\cite{Tong2022VideoMAEMA}.

\subsection{Main Results}
\noindent\textbf{Something-Something V2.}
We first show our \OURMETHOD{}'s performance on SSv2 in~\Cref{table:main_res}, and compare with previous models trained in a supervised manner:
Our \OURMETHOD{} using vanilla ViT-B pretrained on only SSv2 outperforms previous SOTA such as MViTv1~\cite{fan2021multiscale} (+4.4\%) and MViTv2~\cite{fan2021multiscale} (+1.6\%) which uses a heavy hierarchical 3D transformer backbone and pretrains on Kinetics-400. This demonstrates the efficiency and effectiveness of our proposed method compared to supervised 3D transformer backbones. 
\OURMETHOD{} also significantly outperforms previous supervised learning methods that use the same backbone as us: +5.6\% over MotionFormer~\cite{patrick2021keeping} and +12.6\% over TimeSformer~\cite{bertasius2021space}. This demonstrates that the proposed \OURMETHOD{} is effective for spatiotemporal modeling.
% It is worth mentioning that previous spatiotemporal transforfmers are pretrained on Kinetics, the proposed XX is data efficient.

We then compare \OURMETHOD{} with SOTA video MAE methods. With the same backbone and training schema, \OURMETHOD{} achieves $1.3\%$ higher accuracy comparing to VideoMAE~\cite{Tong2022VideoMAEMA}. Note that with 3$\times$ less training time, our \OURMETHOD{} pretrained for 800 epochs is able to achieve nearly the same performance as VideoMAE pretrained for 2400 epochs ($\sim 66\%$ fewer epochs), as illustrated in ~\cref{fig:teaser}. 
% Under the same training budget of 1600 epochs pretraining, the MGM outperforms videoMAE by XX\%.
This demonstrates that motion-guided masking is a more efficient strategy for self-supervised spatiotemporal learning due to learning video saliency. Compared to OmniMAE~\cite{Girdhar2022OmniMAESM} which does masked modeling on both images and video, \OURMETHOD{} achieves +2.6\% improvement. Compared to BEVT~\cite{Wang2022BEVTBP} which also learns from both images and video and additionally uses Swin Transformer~\cite{liu2021swin, liu2022video} -- a heavy 3D Transformer architecture -- \OURMETHOD{} achieves +1.5\% improvement. This is despite the fact that our method only uses video and is pre-trained from scratch. Note that STMAE~\cite{Feichtenhofer2022MaskedAA} only provides results for ViT-Large backbone on SSv2.

We finally compare our \OURMETHOD{} with some most recent works that also utilize motion information for video pretraining.
\OURMETHOD{} outperforms M$^{3}$Video~\cite{Sun2022M3VideoMM} (+$0.4\%$ with the same 400 epochs of pretraining) which reconstructs motion trajectories generated from optical flow. We achieve equivalent performance to MotionMAE~\cite{yang2022self} using 50\% fewer epochs (1600 vs. 2400) even though it utilizes frame difference as an additional reconstruction target which incorporates explicit motion information.
%\OURMETHOD{} also achieves comparable performance to recent work that tries to reconstruct RGB frame difference and visual tokens from dVAE pretrained on OpenImages~\cite{song2022takes}. In contrast, we make no change to the reconstruction target, do not use optical flow, and do not leverage additional pretraining such as OpenImages. 
The results show that motion-guided masking alone is sufficient for improving spatiotemporal learning.
% We simply change the masking strategy to be motion-guided and demonstrate that this alone is sufficient for improving spatiotemporal learning.
It is worth reiterating that our \OURMETHOD{} uses motion vectors for masking guidance, which is already computed during video encoding, and thus directly available during video decoding~\cite{richardson2002video, wu2018compressed, wang2022deformable}, making our method more efficient and scalable comparing to works such as~\cite{Sun2022M3VideoMM, song2022takes, yang2022self} which use optical flow and/or frame difference. 
% This is unlike other approaches~\cite{Sun2022M3VideoMM, song2022takes, yang2022self} which use optical flow and/or frame difference.

\noindent\textbf{Kinetics-400 (K400).} 
We further conduct experiments on K400 and show results in~\Cref{table:main_res}. 
Similar trends are observed. Our \OURMETHOD{} is able to outperform supervised models with the same backbone such as TimeSformer~\cite{bertasius2021space} by +3.7\% and MotionFormer~\cite{patrick2021keeping} by +2.0\%. 
\OURMETHOD{} slightly underperforms some hierarchical 3D transformer backbones such as MViTv2~\cite{li2022mvitv2} by -1.2\% on K400 but MViTv2 is trained with 2$\times$ more frames (32) and uses significantly more FLOPS.
 
Comparing with other MAE based methods, our \OURMETHOD{} consistently outperforms VideoMAE by +$1.4\%$ at 800 epochs of pretraining. \OURMETHOD{} also achieves better performance comparing to other recent works, at no additional cost. Our \OURMETHOD{} outperforms M$^{3}$Video~\cite{Sun2022M3VideoMM} at 400 epochs of pretraining by 0.6\%, while being more efficient as our method does not use frame difference nor optical flow. Compared to OmniMAE~\cite{Girdhar2022OmniMAESM} which uses both images and video, \OURMETHOD{} achieves 0.9\% improvement. Compared to BEVT~\cite{Wang2022BEVTBP}, \OURMETHOD{} achieves 1.1\% improvement.

We note that the performance gain \OURMETHOD{} obtains over other methods on K400 is slightly lower compared to on SSv2. This can be explained by the fact that a great proportion of actions in K400 dataset can be differentiated by the appearance of a few key frames, for example, ``skiing''  \vs ``tennis''. In contrast, recognising actions in SSv2 dataset, \eg ``moving something up'' \vs ``moving something down'', requires the model to understand the subtle differences in motion along temporal dimension. Therefore, we hypothesize that spatiotemporal modeling is less impactful on spatial-heavy datasets such as K400 and the strength of \OURMETHOD{} is not fully exhibited. Motion serves as a appearance-agnostic differentiator between classes.

\subsection{Transfer Learning}
\noindent We next evaluate \OURMETHOD{}'s performance when transferred to smaller datasets: UCF101~\cite{soomro2012ucf101}, HMDB51~\cite{kuehne2011hmdb}, and Diving48~\cite{li2018resound}. 
Performance in transfer learning is commonly used as an indicator of feature quality and representativeness~\cite{He2020MomentumCF}. We use ViT-Base model pretrained on K400 and SSv2 by 800 epochs for all transfer learning experiments.
% Good performance in a transfer learning setting is an indicator of spatiotemporal learning because these small datasets differ widely in appearance. Generalizing suggests that the model understands motion which is agnostic to spatial appearance.

\noindent\textbf{Finetune on downstream tasks.} 
We first present fine-tuning results from \OURMETHOD{} pretrained on unlabeled K400 in ~\Cref{tab:generalization}. 
% and compare it to the VideoMAE baseline~\cite{Tong2022VideoMAEMA} and a supervised baseline of the same backbone of ViT-B~\cite{dosovitskiy2020image} trained on K400 for 200 epochs. 
Our \OURMETHOD{} consistently outperforms VideoMAE with same backbone and pretraining setup on all three datasets (+0.9\% on UCF101, +1.1\% on HMDB51, +4.9\% on Diving48).
This shows that our approach learns more representative semantics and thus generalizes better to small-scale downstream tasks. 
It is worth mentioning that \OURMETHOD{} even outperforms K400 supervised pretraining on HMDB51 (+3.2\%) and on Diving48 (+7.5\%), demonstrating the effectiveness of our proposed method.

Our \OURMETHOD{} is also able to consistently outperform VideoMAE~\cite{Tong2022VideoMAEMA} with pretraining on unlabeled SSv2 in the finetune setting (+1.3\% on UCF101, +1\% on HMDB51, +2.6\% on Diving48). Our \OURMETHOD{} even outperforms SSv2 supervised pretraining by a large margin (+14.1\% on UCF101, +13.4\% on HMDB51, +46.9\% on Diving48). Finetuning \OURMETHOD{} on SSv2 (same model from ~\cref{table:main_res}) does not lead to a large boost, indicating that our model is able to learn the majority of semantics just from unsupervised pretraining, and that unsupervised pretraining is able to make better use of the information in SSv2 than supervised training.  

\begin{table}[t]
\small
\centering
    \subfloat[\textbf{Pretrained on K400} and finetuned on downstream tasks.]
    {
    \scalebox{0.95}{
       \begin{tabular}{lccc}
            \toprule
		  & {{UCF}} & {{HMDB}} & {{Diving48}}  \\
            \midrule
            Supervised (K400) $\dagger$                             & 95.1     & 71.4     & 62.7    \\
            VideoMAE~\cite{Tong2022VideoMAEMA}      & 93.3  & 73.5  & 65.3 \\
            \textbf{\OURMETHOD{}}                   & 94.2  & 74.6  & 70.2  \\
            + K400 Finetune (\cref{table:main_res}) & 97.7  & 81.0  & 82.6  \\
            \bottomrule 		
        \end{tabular}
        \label{tab:finetune_k400}
        }
    }\hfill
    \subfloat[\textbf{Pretrained on SSv2} and finetuned on downstream tasks.]
    {
    \scalebox{0.95}{
       \begin{tabular}{lccc}
            \toprule
		  & {{UCF}} & {{HMDB}} & {{Diving48}}  \\
            \midrule
            Supervised (SSv2) $\dagger$                         & 77.8 & 56.3 & 36.2  \\
            VideoMAE~\cite{Tong2022VideoMAEMA}      & 90.6 & 68.7 & 80.5 \\
            \textbf{\OURMETHOD{}}                   & 91.9 & 69.7 & 83.1   \\
            + SSv2 Finetune (~\cref{table:main_res}) & 93.2 & 74.7 & 83.4   \\
            \bottomrule 		
        \end{tabular}
        \label{tab:finetune_SS}
        }
    }\hfill
    \subfloat[\textbf{Pretrained on K400} and evaluated with linear probing on downstream tasks.]
    {
    \scalebox{0.95}{
         \begin{tabular}{lccc}
            \toprule
		  & {{UCF}} & {{HMDB}} & {{Diving48}}   \\
            \midrule
            Supervised (K400) $\dagger$                         & 94.1      & 65.4      & 25.2             \\
            VideoMAE~\cite{Tong2022VideoMAEMA}      & 70.5      & 45.4      & 10.0  \\
            \textbf{\OURMETHOD{}}                   & 77.5      & 49.2      & 14.2   \\
            \bottomrule	\end{tabular}
        }
        \label{tab:linear_probe_k400}
    }\hfill
    \subfloat[\textbf{Pretrained on SSv2} and evaluated with linear probing on downstream tasks.]
    {
    \scalebox{0.95}{
         \begin{tabular}{lccc}
            \toprule
		  & {{UCF}} & {{HMDB}} & {{Diving48}}  \\
            \midrule
            Supervised (SSv2) $\dagger$                         & 72.9 & 51.8 & 13.2    \\
            VideoMAE~\cite{Tong2022VideoMAEMA}      & 65.3 & 41.2 & 10.7   \\
            \textbf{\OURMETHOD{}}                   & 69.1 & 44.8 & 10.2   \\
            \bottomrule	\end{tabular}
        }
        \label{tab:linear_probe_SS}
    }
    \caption{\OURMETHOD{} generalizes well to various downstream datasets (UCF101, HMDB51 and Diving48) on two downstream tasks (finetune, linear-probe). We test VideoMAE and \OURMETHOD{} pretrained on Kinetics-400 (K400) for 800 epochs. $\dagger$ The supervised baseline is ViT-B trained from scratch on K400 and SSv2 using the recipe from MViT~\cite{fan2021multiscale, li2022mvitv2}.}
    \label{tab:generalization}
	% \vspace{-5mm}
\end{table}

\noindent\textbf{Linear Probe on downstream tasks.} 
Previous work~\cite{bao2021beit, he2022masked, Feichtenhofer2022MaskedAA} argues that generative approaches such as MAE generally perform worse on linear probing tasks as there is a larger gap between the reconstruction task and downstream evaluation compared to other pretraining methods. Nevertheless, we compare linear probe performance between \OURMETHOD{} and VideoMAE~\cite{Tong2022VideoMAEMA} and show that our method achieves significantly better performance. With K400 pretraining, we outperform VideoMAE by +7\% on UCF101, +3.8\% on HMDB51, and +4.2\% on Diving48. With SSv2 pretraining, we outperform by +3.8\% on UCF101, +3.6\% on HMDB51, and underperform by -0.5\% on Diving48. This demonstrates that motion-modeling helps reduce the gap between the reconstruction task and downstream tasks, as linear probe only trains a linear clasifier on top of features extracted from the frozen backbone; linear layers cannot learn new semantics.

We note that there is a large gap of over 10\% with K400 supervised pretraining on linear probe. This is a general deficiency of generative methods such as MAE-based pretraining~\cite{Feichtenhofer2022MaskedAA} and we leave further exploration of this problem to future works. One way to address this would be to combine MAE with discriminative methods such as contrastive learning to obtain a balanced representation.

\begin{table}[t]
\footnotesize
\centering
    % \subfloat[\OURMETHOD{}]
    % {
    \scalebox{0.95}{
         \begin{tabular}{lcc}
            \toprule
		  & {{UCF14 $\rightarrow$ HMDB7}} & {{HMDB14 $\rightarrow$ UCF7}}  \\
            \midrule
            VideoMAE~\cite{Tong2022VideoMAEMA}      & 55.7  & 69.6  \\
            \textbf{\OURMETHOD{}}                   & 57.6  & 71.4  \\
            + K400 Finetune (\cref{table:main_res}) &  79.1    &   96.2  \\
            \bottomrule	\end{tabular}
        }
        \label{tab:domain_adaptation_k400}
    \caption{\OURMETHOD{} works well when trained on one dataset and evaluated on another dataset for an overlapping set of classes. The model sees no samples from test dataset, indicating motion understanding as motion is agnostic to spatial appearance. We tested VideoMAE and \OURMETHOD{} pretrained on Kinetics-400 (K400) for 800 epochs.}
    \label{tab:domain_adaptation}
	% \vspace{-5mm}
\end{table}

\noindent\textbf{Domain adaptation.} 
We further evaluate \OURMETHOD{} on partial domain adaptation tasks following~\cite{xu2021partial}, in which the labels for the target dataset are a subset of the labels for the source dataset. The model is finetuned on the source dataset and directly evaluated on the target dataset without seeing any samples from the target dataset. This is more challenging than conventional domain adaptation as the model may incorrectly label samples in the target dataset as the labels from the source dataset which are not present in the target dataset (``negative transfer'')~\cite{xu2021partial}. 
Previous work has showed that optical flow is useful for action recognition because it is invariant to appearance~\cite{sevilla2018integration}. Several works such as~\cite{choi2020shuffle, sahoo2021contrast} improve unsupervised domain adaptation in video by reducing background and appearance bias to capture the essence of an action via shuffling and mixup respectively. By similar reasoning, domain adaptation is a good evaluation protocol to assess motion learning as the source and target datasets have very different visual appearances, and motion serves as a common link between classes that is agnostic to spatial appearances.
% . Motion is the common link between the shared classes labels in the source and target datasets, and motion is agnostic of spatial appearances.

We use the training sets provided by~\cite{xu2021partial} for UCF14 $\rightarrow$ HMDB7 and HMDB14 $\rightarrow$ UCF7.
% and evaluate our \OURMETHOD{} pretrained on K400 and SSv2 for 800 epochs. 
Our \OURMETHOD{} consistently outperform VideoMAE with K400 pretraining in both settings (+1.9\% on HMDB7 and +1.8\% on UCF7) indicating that \OURMETHOD{} can also outperform in a domain adaptation setting. While there is a significant gap of over 20\% between~\Cref{tab:domain_adaptation} and~\Cref{tab:generalization}, finetuning our model on K400 labels bridges the gap. This suggests that our model has learned video saliency and can perform well without any large-scale label supervision, but additional label supervision is helpful for challenging settings such as domain adaptation where spatial information is very different and difficult to learn from limited samples. There are only $\sim$1000 training samples in each setting.

\begin{figure}[t]
\begin{subfigure}[t]{.49\linewidth}
    \centering
    \includegraphics[width=0.99\textwidth]{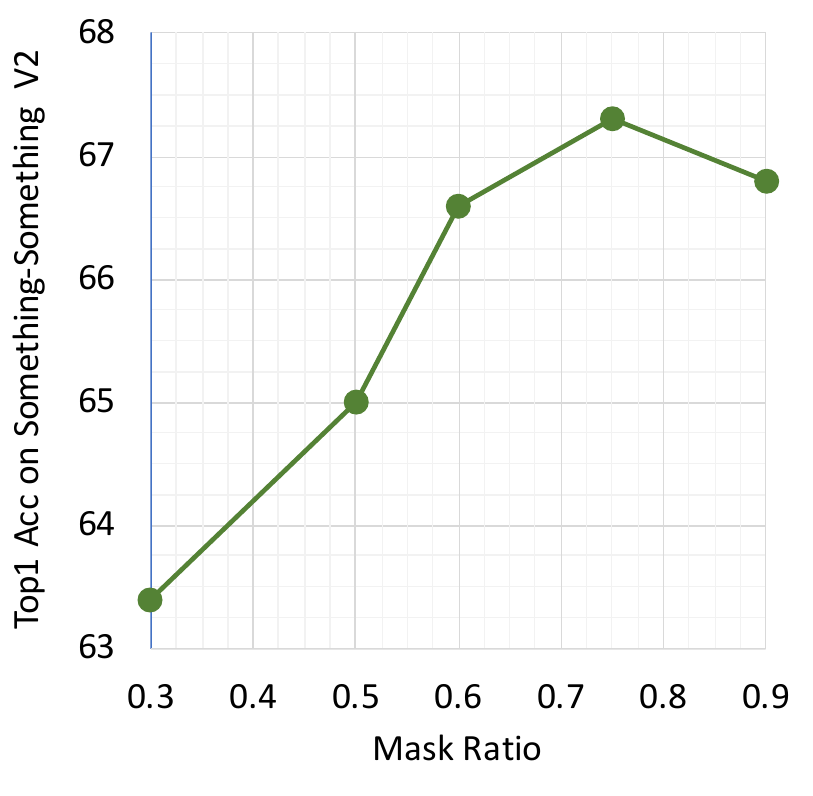}
    \caption{More mask ratios.}
    \label{fig:mask_ratio_ablation}
\end{subfigure}
\begin{subfigure}[t]{.49\linewidth}
    \centering
    \includegraphics[width=0.99\textwidth]{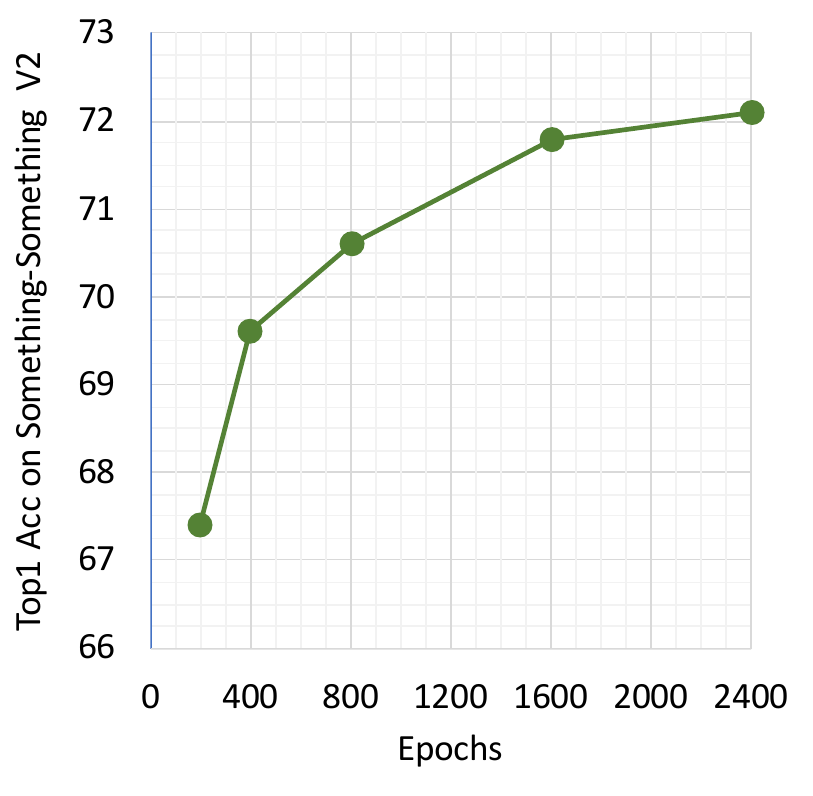}
    \caption{Pretrain duration in epochs.} % Repeated augmentation=2~\cite{hoffer2020augment}
    \label{fig:epoch_ablation}
\end{subfigure}
\caption{Ablations on mask ratio~\ref{fig:mask_ratio_ablation} and pretraining duration~\ref{fig:epoch_ablation}. Mask ratio 0.75 is optimal and performance does not saturate with longer pretraining.}
\end{figure}

\begin{table}[t]
\small
\centering
    \subfloat[Backbone generalization.]
    {
    \scalebox{0.9}{
        \begin{tabularx}{0.21\textwidth}{lc}
        \toprule
        {\textit{backbone}}  &  {\textit{SSv2}} \\
            \midrule
               ViT-S               & 61.1 \\
               \textbf{ViT-B}               & \textbf{67.3} \\
               TimesFormer~\cite{bertasius2021space}            & 64.7  \\
        \bottomrule	
        \end{tabularx}
        }
        \label{tab:backbone}
    }\hfill
    \subfloat[Source of motion info.]
    {
    \scalebox{0.9}{
        \begin{tabularx}{0.21\textwidth}{lc}
        \toprule
        {\textit{guidance}}  &  {\textit{SSv2}} \\ % &  {\textit{K400}}\\
            \midrule
               optical flow           & 66.0 \\ % & 78.6 \\
               \textbf{motion vector}           & \textbf{67.3} \\ % & 80.3* \\
        \bottomrule	
        \end{tabularx}
        }
        \label{tab:motion_source}
    }\hfill
    \subfloat[Mask jitter.]
    {
    \scalebox{0.9}{
        \begin{tabularx}{0.17\textwidth}{cc}
        \toprule
        {\textit{aspect ratio}}  &  {\textit{SSv2}} \\ % &  {\textit{K400}}\\
            \midrule
               no           & 67.1 \\ % & 78.6 \\
               \textbf{yes}           & \textbf{67.3} \\ % & 80.3* \\
        \bottomrule	
        \end{tabularx}
        }
        \label{tab:motion_jitter}
    }\hfill
    \subfloat[Dataset scale generalization.]
    {
    \scalebox{0.9}{
        \begin{tabularx}{0.29\textwidth}{lcc}
        \toprule
        {\textit{pretrain data}}  & {\textit{samples}} &  {\textit{K400}} \\ % &  {\textit{K400}}\\
            \midrule
               Mini-K200~\cite{xie2018rethinking} & 80K & 78.2 \\ % & 78.6 \\
               K400~\cite{kay2017kinetics}        & 240K & 78.6 \\ % & 80.3* \\
        \bottomrule	
        \end{tabularx}
        }
        \label{tab:data_scale}
    }\hfill
    \subfloat[Masking strategies differentiated by initial position, temporal propagation, and spatial continuity. Spatiotemporally continuous masks with motion guidance perform the best.]
    {
    \scalebox{0.78}{
        % \begin{tabularx}{0.26\textwidth}{lc}
        % \toprule
    
        % \bottomrule	
        % \end{tabularx}
        \begin{tabular}{lccc|cc|c}
         & \multicolumn{3}{c}{\textit{\textbf{temporal propagation}}} & \multicolumn{2}{c}{\textit{\textbf{spatial continuity}}} &   \\ 
    		\cmidrule(lr){2-4} \cmidrule(lr){5-6} 
        \textbf{Mask} & \textbf{static} & \textbf{simulated} & \textbf{motion} &  \textbf{sparse} & \textbf{dense} & Acc \\
        \hline
        Random Block & \checkmark{} & & & & \checkmark{} & 62.9 \\
        SMM (Sparse) & & \checkmark{} & & \checkmark{} & & 64.6 \\
        Random Tube & \checkmark{} & & & \checkmark{} & & 65.0 \\
        \OURMETHOD{} (Sparse) & & & \checkmark{} & \checkmark{} & & 65.1 \\
        % Random + MGM (sparse) & & & & & & & & 65.8 \\
        SMM (Dense) & & \checkmark{} & & & \checkmark{} & 65.9 \\
        % SMM (dense) + \OURMETHOD{} (sparse) & & & & & & & & 66.2 \\
        \textbf{\OURMETHOD{}} (Dense) & & & \textbf{\checkmark{}} & & \textbf{\checkmark{}} & \textbf{67.3} \\
      \end{tabular}
        }
        \label{tab:mask_type}
    }\hfill
   \caption{Ablations on backbone~\ref{tab:backbone}, source of motion guidance~\ref{tab:motion_source}, mask jitter~\ref{tab:motion_jitter}, pretrain data scale~\ref{tab:data_scale}, and mask type~\ref{tab:mask_type}. All ablations pretrain for 200 epochs on SSv2 unless otherwise indicated.}
	\label{tab:ablation}
	% \vspace{-5mm}
\end{table}

\subsection{Ablations}
\noindent We perform ablations on the Something-Something V2 dataset, with \OURMETHOD{} with ViT-B backbone pretrained for 200 epochs using our motion-guided masking, unless otherwise specified.

\noindent\textbf{Impact of masking ratio. }
~\Cref{fig:mask_ratio_ablation} studies the impact of different masking ratio on SSv2. We noticed that 0.75 masking ratio works best for the proposed \OURMETHOD{}, which is lower than that of other works~\cite{Tong2022VideoMAEMA, Feichtenhofer2022MaskedAA}.
This is because motion-guided masking leads to a more ``challenging'' reconstruction task compared to random masking, since the masked regions of videos are most informative and therefore cannot be easily inferred from retained regions.

\noindent\textbf{Impact of pretraining length.}
We show the impact of different pretraining epochs on  SSv2 in ~\Cref{fig:epoch_ablation}. The proposed \OURMETHOD{} consistently improves with more training and we achieve state-of-the-art performance on SSv2 with 2400 epochs of pretraining. Performance does not seem to saturate with longer pretraining.

\noindent\textbf{Generalization to different backbones. }
\OURMETHOD{} also generalizes well to backbones of different sizes and types (~\Cref{tab:backbone}). We test vanilla ViT-S~\cite{dosovitskiy2020image} and  ViT-B~\cite{dosovitskiy2020image}, and TimeSformer~\cite{bertasius2021space} with joint space-time attention. 

Note that our TimeSformer ablation outperforms the supervised result reported by~\cite{bertasius2021space} of 59.5\% on SSv2 which uses IN21K pretraining. In contrast, we do not use any extra data.
ViT-S's lower performance may be explained by its low model capacity which may not be sufficient for learning spatiotemporal dynamics. As other methods do not use ViT-S, we cannot compare our result to other methods. We thus use ViT-B to compare fairly against other MAE works which use ViT-B. 

\noindent \textbf{Impact of motion source.}
In Table~\ref{tab:motion_source} we compare the performance of \OURMETHOD{} when using optical flow vs. motion vectors as the source of motion guidance. Even though optical flow is more fine-grained and precise than motion vectors, performance of \OURMETHOD{} drops when the mask is guided by optical flow. This could be related to the fact that optical flow is calculated on a per-pixel basis whereas motion vectors are computed in a block-wise manner and we utilize a block-wise mask strategy. We thus conveniently choose to use motion vector as it is also directly obtainable from compressed video. We extract optical flow using RAFT-Small~\cite{teed2020raft}. On the SSv2 dataset, reading motion vectors is roughly 30$\times$ faster than computing optical flow (80 milliseconds vs. 2.5 seconds per video on average).

\noindent \textbf{Impact of mask aspect ratio jitter.}
In Table~\ref{tab:motion_jitter} we ablate the use of aspect ratio jittering from ~\cref{equ:mgm_masking}. We see that aspect ratio jittering gives a small 0.2\% boost in performance but is not a critical component of \OURMETHOD{}.

\noindent \textbf{Impact of pretrain dataset scale.}
In Table~\ref{tab:data_scale} we ablate the amount of pretraining data used to see if \OURMETHOD{} can generalize to different dataset scales. We pretrain \OURMETHOD{} on MiniKinetics-200~\cite{xie2018rethinking}  and Kinetics-400 (K400)~\cite{kay2017kinetics} which contain $\sim$80K videos and $\sim$240K videos respectively for 200 epochs. MiniKinetics-200 contains videos belonging to a subset of 200 of the most common classes in K400. We then evaluate top-1 accuracy on K400. We find that there is a small 0.4\% drop in performance when pretraining on the smaller MiniKinetics-200. We expect performance to further improve when pretraining on datasets larger than K400.

\begin{figure}[th]
    \centering
    \includegraphics[width=.99\linewidth]{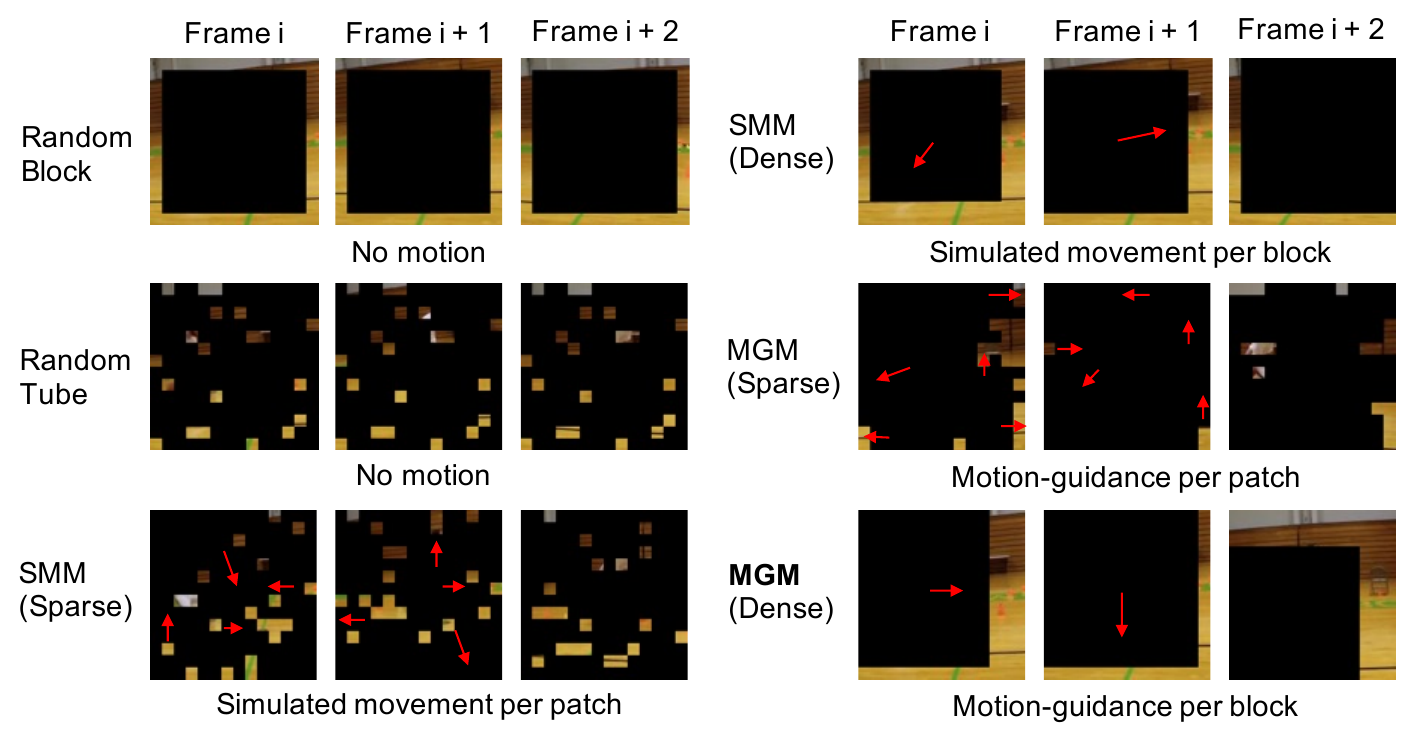}
    \caption{Visualization of different mask algorithms.}
    \label{fig:all_mask_visualization}
\end{figure}

\noindent\textbf{Effectiveness of 3D motion masking.}
% First argue dense masking is better whether it's SMM or MGM
% Then compare MGM to SMM
Table~\ref{tab:mask_type} studies different masking strategies which are broken down into categories based on degree of motion guidance and spatial continuity. We first compare dense to sparse masking, where sparse means that the mask may be spatially discontinuous. The dense variant of SMM and \OURMETHOD{} outperform the respective sparse counterpart; dense SMM outperforms sparse SMM by 1.3\% and dense \OURMETHOD{} outperforms sparse \OURMETHOD{} by 2.2\%. This suggests the importance of maintaining spatial continuity in the masked reconstruction. Next, we compare \OURMETHOD{} to SMM. Sparse \OURMETHOD{} outperforms sparse SMM by 0.5\% and dense \OURMETHOD{} outperforms dense SMM by 1.4\%. When motion is completely removed, performance is the worst even when the mask is spatially continuous (static block gets 62.9\%). Dense \OURMETHOD{} outperforms all masking strategies and specifically the random masking baseline by 2.3\%. All masking strategies except block and sparse SMM outperform random masking. Random masking is spatiotemporally discontinuous. This validates our intuition that forming continuous 3D masks which model the true motion information within video is useful. We thus pick dense \OURMETHOD{} for all other experiments and refer to ``\OURMETHOD{}'' as the dense variant by default. For a visualization of these masking strategies, see~\Cref{fig:all_mask_visualization}.

\subsection{Visualization}
\begin{figure}[t]
    \centering
    \includegraphics[width=0.45\textwidth]{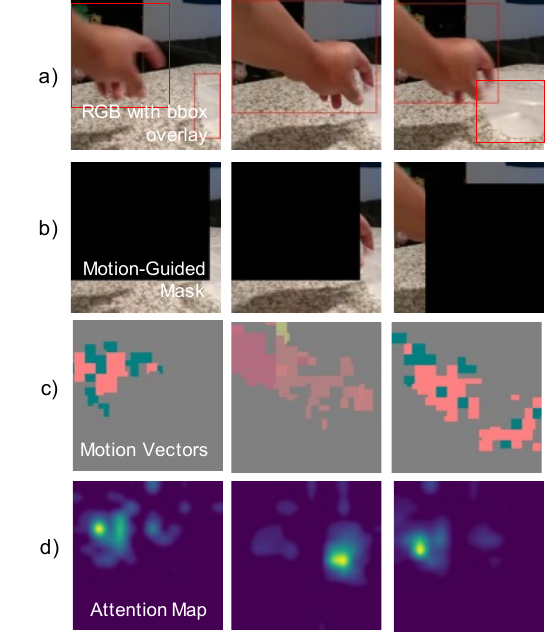}
    \caption{a) RGB frames with bounding boxes for visualization. b) Our \OURMETHOD{} masks continuously cover the moving hand. c) Motion vector maps. d) Encoder attention map using center patch of center frame as query.}
    \label{fig:qualitative_visualization}
    % \vspace{-5mm}
\end{figure}

% To gain more insight of the proposed model, 
\noindent In~\Cref{fig:qualitative_visualization}, we provide visualizations from two perspectives using \OURMETHOD{} pretrained on K400. Note that visualizations are subjective and should not be used as a formal explanation for model behavior. Our intention is to provide additional insight into the model to complement our quantitative results. We first visualize the RGB frames with bounding boxes, motion-guided masks, and motion vectors. We see that the motion-guided masks overlap with the majority of spatiotemporally salient regions. Second, we visualize the attention map using the center patch of the center frame as a query. We observe that the model mostly attends to the salient regions. For more visualizations, see the supplementary.

\section{Discussion and Conclusion}
\label{sec:Conclusion}
% We demonstrated that motion-guided masking leads to better learning as it makes video reconstruction a more challenging task that cannot be solved by copying adjacent video patches, but rather requiring understanding spatiotemporal information. Our method leverages motion vectors which come for free when decoding H.264 videos. We believe that the computational efficiency of MAE combined with our scalable approach of leveraging motion vectors will enable the use of MAE for long-form video in the near future.

\noindent\textbf{Limitations.} Our method efficiently leverages motion cues from compressed video format. However, \OURMETHOD{}'s effectiveness on the video benchmarks that are not human centric or have tremendous camera motion has yet to be examined. To the best of our knowledge, we did not find such a video benchmark, so we will leave this to future work. 

\noindent\textbf{Potential negative impact.} 
As our model is trained on primarily Internet video datasets, our model will learn the bias inherent to that data. There could be unintended consequences and we advocate for complying with the law.

\noindent\textbf{Conclusion.} This paper introduces motion-guided masking (\OURMETHOD{}), an algorithm which produces motion-aware 3D masks to improve spatiotemporal learning. Our use of motion information is much more efficient than previous optical flow as we leverage motion vectors that naturally exist in the video codec. We achieve new state-of-the-art or comparable performance on two challenging large-scale video benchmarks and achieve previous state-of-the-art results with 50\% less pretraining time, making our method more efficient. We also achieve better generalization than previous video MAE works on three small-scale datasets. We believe our method has the potential to enable more efficient video training.

% an improved masking strategy for learning spatiotemporal dynamics from video that leverages motion information rather than assuming the salient parts of video are uniformly distributed. Our use of motion vectors allows \OURMETHOD{} to obtain motion information from decoded videos with no additional computation, making our method efficient and scalable. \OURMETHOD{} achieves state-of-the-art or comparable performance on multiple large and small video benchmarks.

% \xl{I would refine this section, but I will work with David on it tomorrow.}

{\small
\bibliographystyle{ieee_fullname}
\bibliography{main}
}

\newpage

\appendix
\section{Additional Hyperparameters}
We mostly follow the same hyperparameters as~\cite{Tong2022VideoMAEMA}. ~\Cref{table:hyperparameters_pretraining} and \Cref{table:hyperparameters_finetuning} show the configurations for pretraining and finetuning.

\begin{table}[!th]
\small
\begin{center}
    \begin{tabular}{l|cc}
    config & SSv2 & K400 \\
    \hline
    optimizer & \multicolumn{2}{c}{AdamW} \\
    base learning rate$^{\dagger}$ & \multicolumn{2}{c}{1.5e-4} \\
    weight decay & \multicolumn{2}{c}{0.05} \\
    optimizer momentum & \multicolumn{2}{c}{$\beta_1, \beta_2=0.9, 0.95$} \\
    batch size & \multicolumn{2}{c}{512} \\
    learning rate schedule & \multicolumn{2}{c}{cosine decay~\cite{loshchilov2016sgdr}} \\
    warmup epochs & \multicolumn{2}{c}{40} \\
    flip augmentation & no & yes \\
    augmentation & \multicolumn{2}{c}{MultiScaleCrop}
  \end{tabular}
\end{center}
\vspace{-2mm}
\caption{Pretraining hyperparameters. $^{\dagger}:$ we follow the linear LR scaling rule. $lr=base\_lr \times batch\_size / 256$.}
\label{table:hyperparameters_pretraining}
\end{table}

\begin{table}[!th]
\small
\begin{center}
    \begin{tabular}{l|cccccc}
    config & SSv2 & K400 & AVA & UCF101 & HMDB51 & Diving48 \\
    \hline
    optimizer & \multicolumn{6}{c}{AdamW} \\
    base learning rate & 5e-4 & 1e-3 & 2.5e-4 & 5e-4 & 5e-4 & 5e-4 \\
    weight decay & \multicolumn{6}{c}{0.05} \\
    optimizer momentum & \multicolumn{6}{c}{$\beta_1, \beta_2=0.9, 0.999$} \\
    layer-wise lr decay & 0.75~\cite{bao2021beit} & 0.75 & 0.7 & 0.75 & 0.7 & 0.7 \\
    batch size & \multicolumn{6}{c}{128} \\
    learning rate schedule & \multicolumn{6}{c}{cosine decay} \\
    repeated augmentation & 2~\cite{hoffer2020augment} & 2 & 1 & 2 & 2 & 2 \\
    warmup epochs & 5 & 5 & 5 & 0 & 0 & 0 \\
    total epochs & 30 & 75 & 30 & 100 & 100 & 100 \\
    flip augmentation & no & yes & yes &  yes & yes & yes \\
    drop path & 0.1 & 0.1 & 0.2 & 0.2 & 0.2 & 0.2
  \end{tabular}
\end{center}
\caption{Finetuning hyperparameters.}
\label{table:hyperparameters_finetuning}
\end{table}

\section{Additional Visualizations}
\begin{figure*}[!th]
    \centering
    \includegraphics[width=0.9\textwidth]{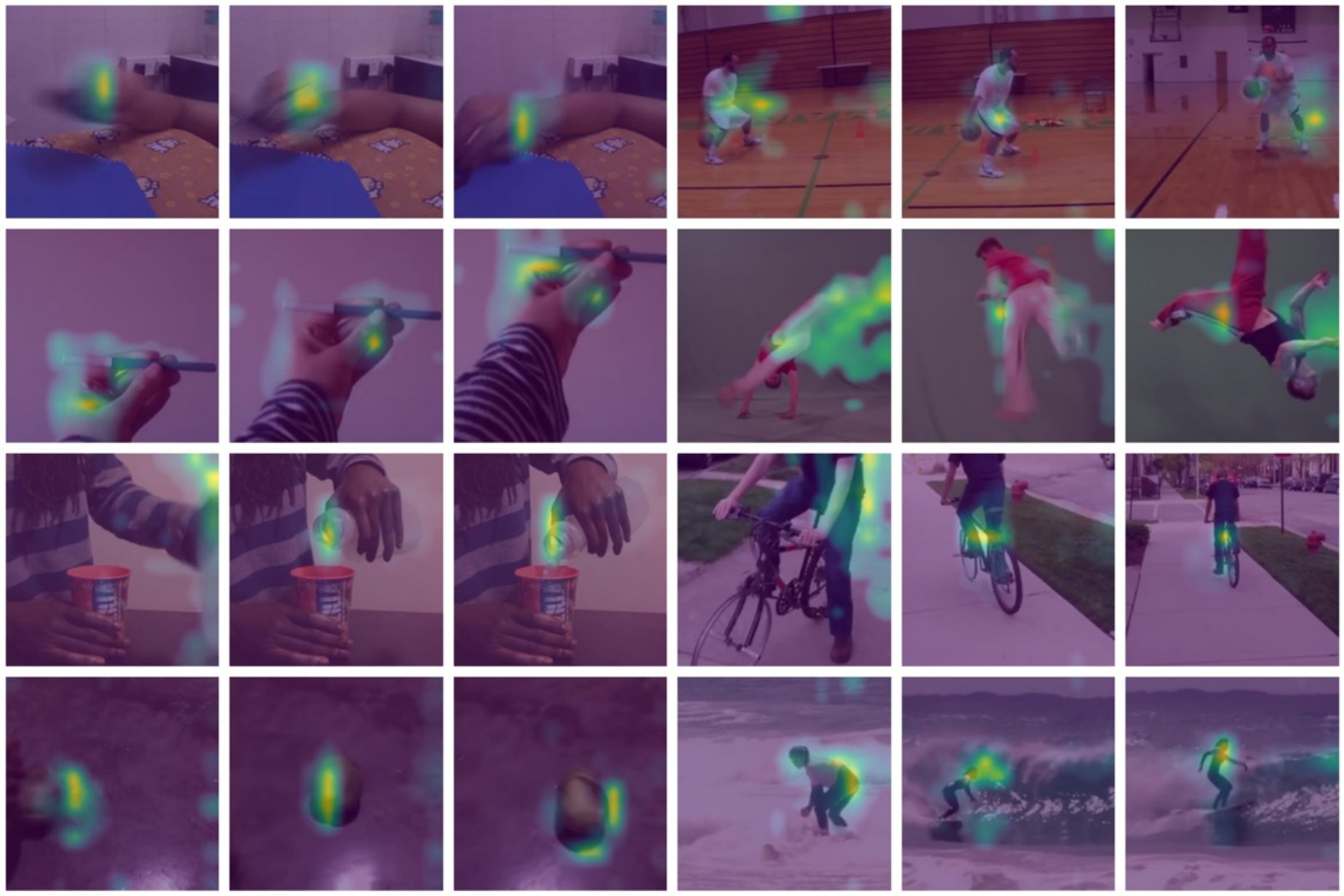}
    \caption{Encoder attention visualization overlaid to RGB frames where the query is the center patch of the center frame. Our \OURMETHOD{} attends mostly to the salient regions of motion across frames.}
    \label{fig:qualitative_visualization_attention}
\end{figure*}

\begin{figure*}[!th]
    \centering
    \includegraphics[width=0.48\textwidth]{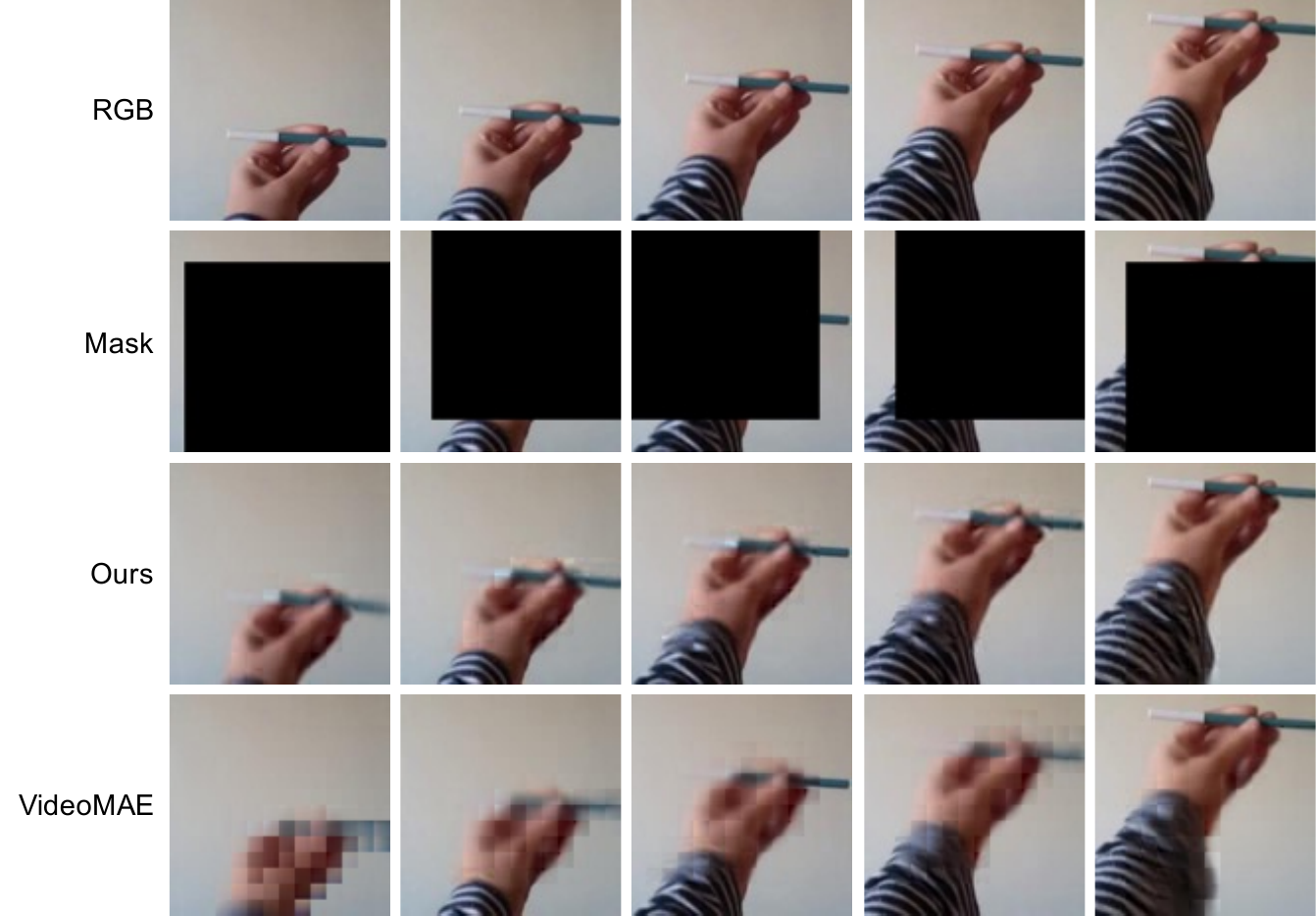}
    \includegraphics[width=0.48\textwidth]{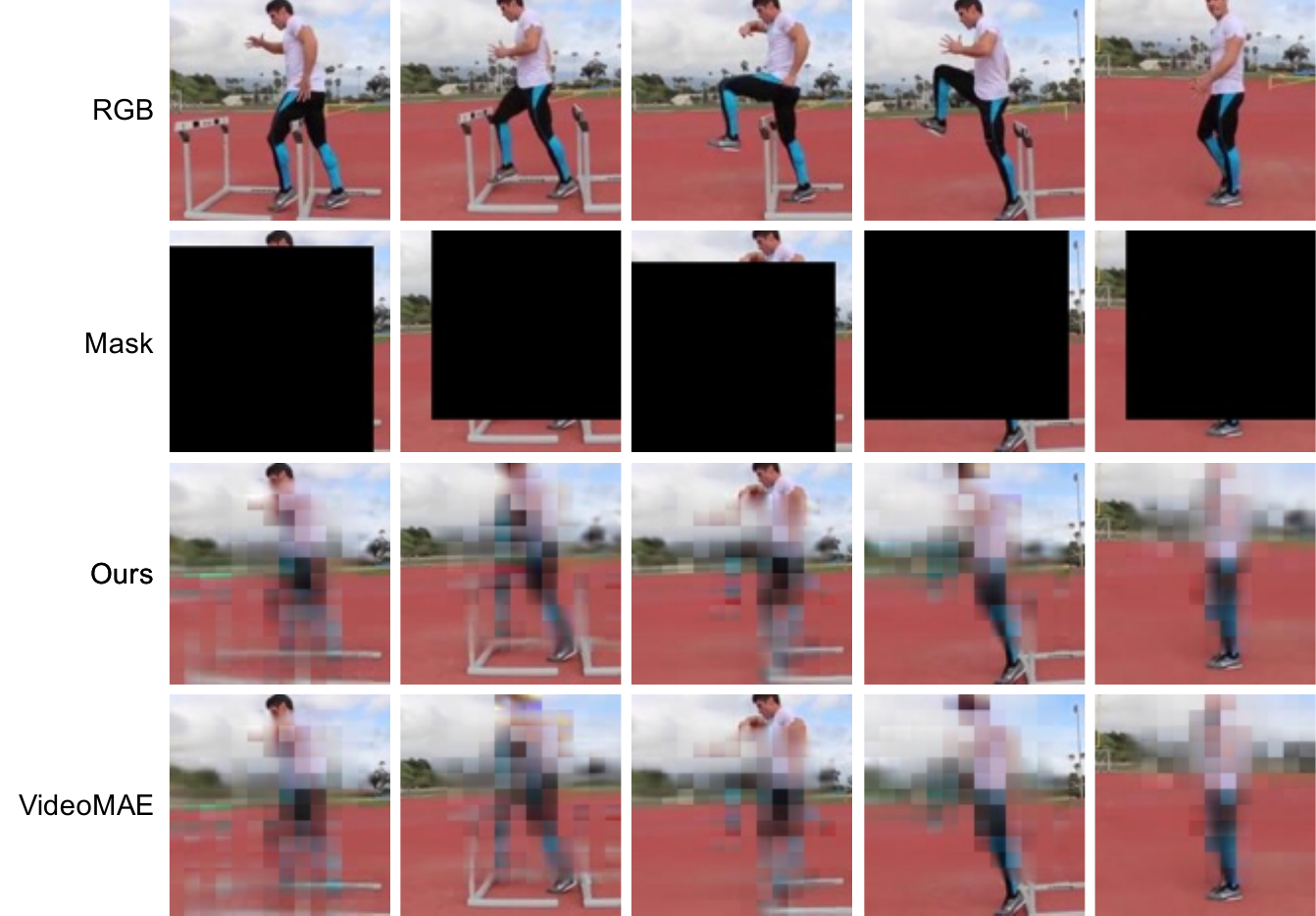}
    \includegraphics[width=0.48\textwidth]{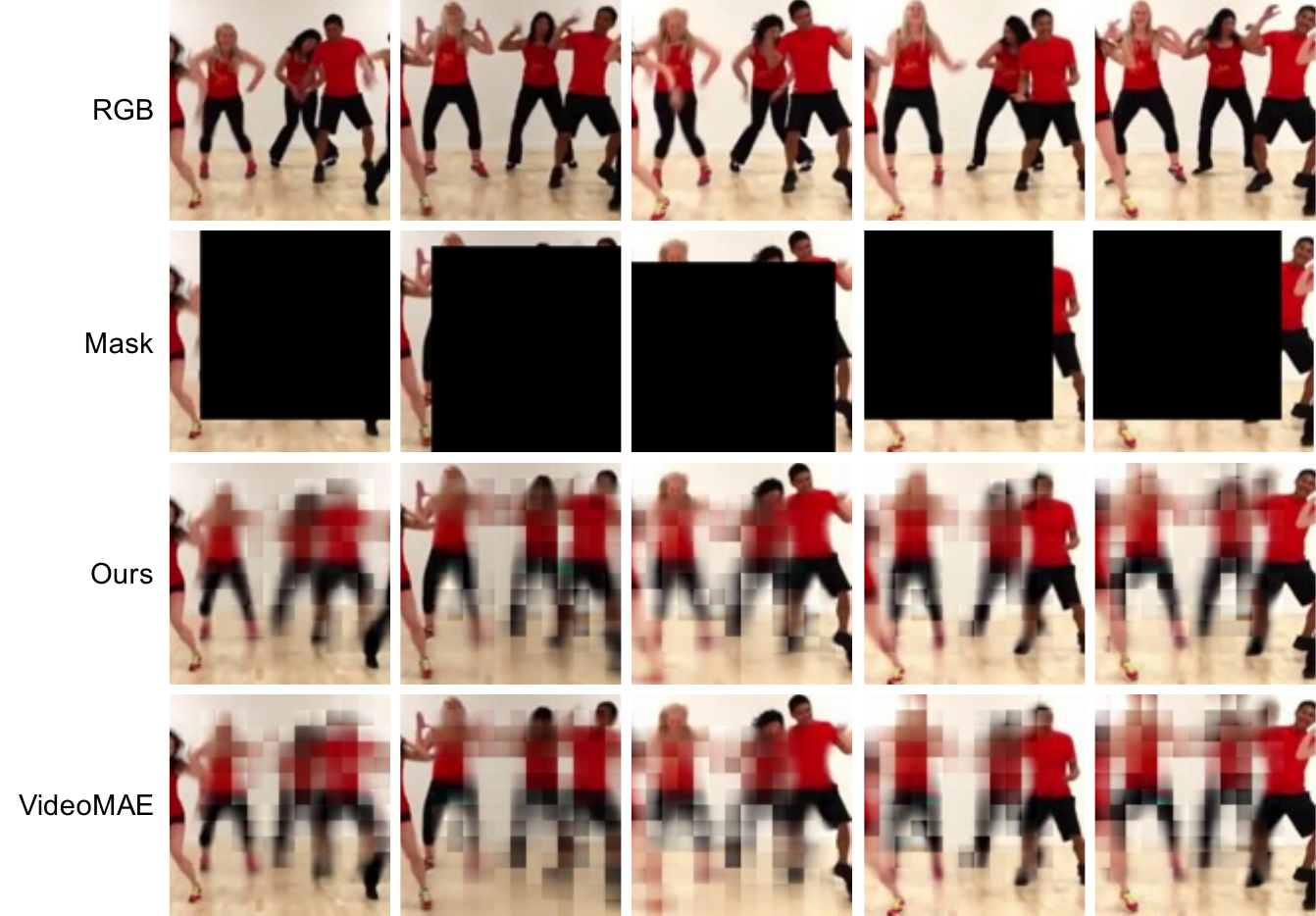}
    \includegraphics[width=0.48\textwidth]{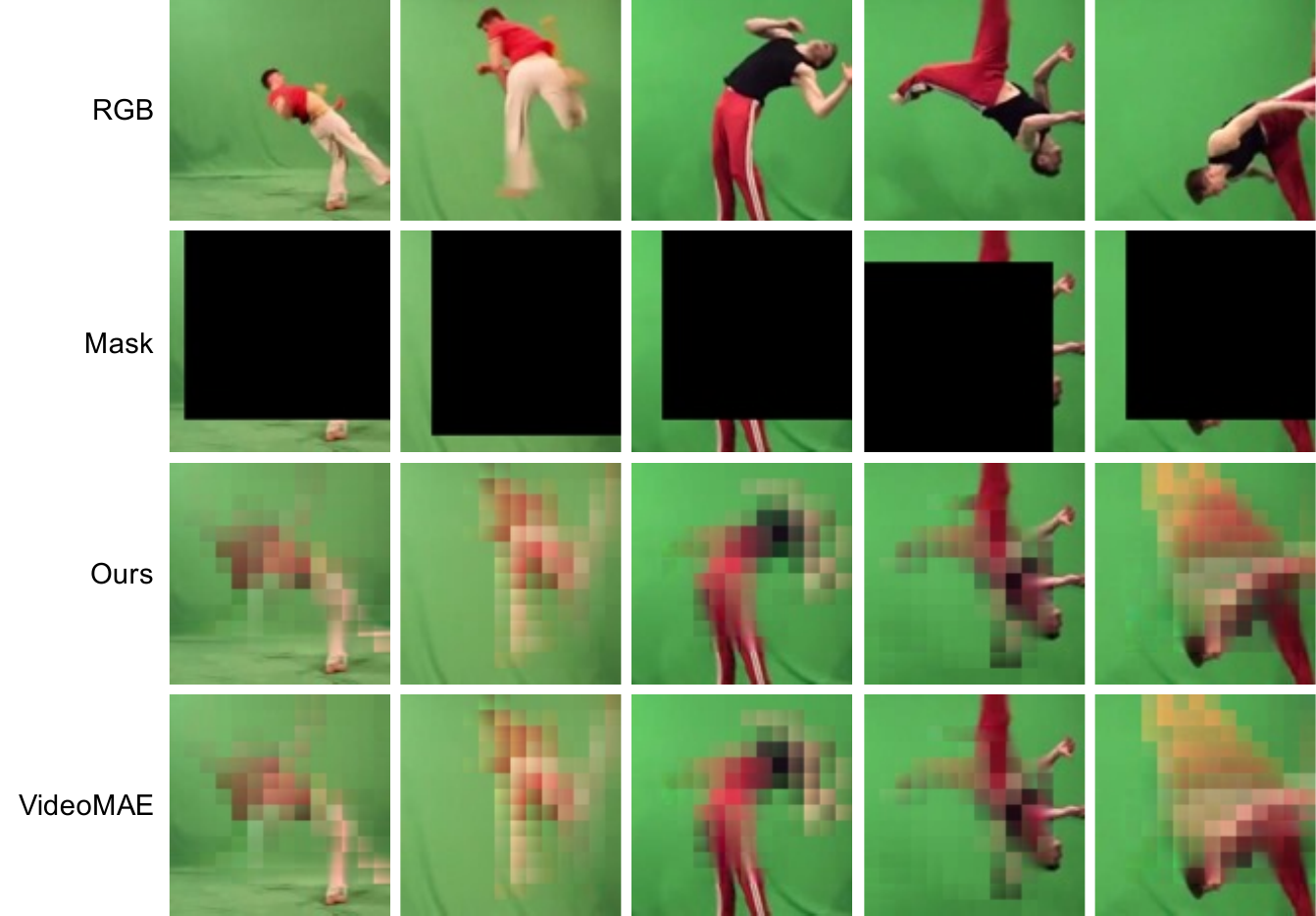}
    \caption{Our \OURMETHOD{} achieves similar if not better reconstruction quality as VideoMAE despite using motion-guided masking which makes the reconstruction task more difficult. The masked regions of videos are most informative and therefore cannot be easily inferred from non-masked regions. \OURMETHOD{} is forced to learn spatiotemporal semantics throughout the video to reconstruct the spatiotemporally continuous motion-guided masked regions.}
    \label{fig:qualitative_visualization_reconstruction}
\end{figure*}

In \Cref{fig:qualitative_visualization_attention} we show more attention visualizations for our \OURMETHOD{} overlaid on top of the RGB frames. \OURMETHOD{} seems to attend mostly to the salient video regions. In \Cref{fig:qualitative_visualization_reconstruction} we visualize more masks and RGB reconstructions for \OURMETHOD{} and VideoMAE. Despite \OURMETHOD{} being forced to solve a more challenging reconstruction task, it is able to achieve similar if not better reconstruction quality than VideoMAE. Again, we emphasize that visualizations are not intended to provide a formal explanation for model behavior. Our intention is to provide additional insights into the model to complement our quantitative results.

% more explanation
% follows direction of motion
% better linear probe generalization
% try blurring the map
% try multiple queries

% - ViT-S, ViT-B, TimeSformer ablation on K400
% - SSv2 K400 1600 epochs
% - More types of mask strategies with visualizations
% reconstruction experiment with fixed backbone

\end{document}